\documentclass[lettersize,journal]{IEEEtran}
\usepackage{xspace}
\usepackage[normalem]{ulem}

\newcommand{\ccreid}{CC-ReID\xspace}
\newcommand{\reid}{Re-ID\xspace}

\newcommand\etal{\textit{et al.}\xspace}

\newcommand\modelnameshort{IDNet\xspace}
\newcommand\ie{i.e.\xspace}

\expandafter\let\csname equation*\endcsname\relax
\expandafter\let\csname endequation*\endcsname\relax
\usepackage{amssymb,amsthm,amsmath}
\usepackage{pifont}
\usepackage{utfsym}
\usepackage{bbding}
\usepackage{wasysym}
\usepackage{bbm}
\usepackage{algorithmic}
\usepackage{algorithm}
\usepackage{array}
\usepackage[caption=false,font=normalsize,labelfont=sf,textfont=sf]{subfig}
\usepackage{textcomp}
\usepackage{stfloats}
\usepackage{url}
\usepackage{verbatim}
\usepackage{graphicx}
\usepackage{cite}
\usepackage{multirow}
\usepackage{ulem}
\definecolor{linkcolor}{HTML}{0071bc}

\usepackage[normalem]{ulem}
\usepackage[colorlinks,
            allcolors=black,
            linkcolor=linkcolor,
            citecolor=green,
            ]{hyperref}

\begin{document}

\newcommand{\bing}[1]{\textcolor{blue}{#1}}
\newcommand{\bingdel}[1]{\textcolor{red}{\sout{#1}}}
\newcommand{\bingadd}[1]{\textcolor{green}{[#1]}}

\newcommand{\bingmodify}[2]{\textcolor{red}{\sout{#1}} \textcolor{green}{[#2]}}
\title{Identity-aware Dual-constraint Network for Cloth-Changing Person Re-identification}

\author{Peini Guo, Mengyuan Liu, Hong Liu, Ruijia Fan, Guoquan Wang, Bin He
\thanks{P. Guo, M. Liu, H. Liu, R. Fan and G. Wang are with the Key Laboratory of Machine Perception, {Shenzhen Graduate School, Peking University}, China (e-mail: {\url{guopeini@stu.pku.edu.cn}, \url{liumengyuan@pku.edu.cn}, \url{hongliu@pku.edu.cn}, \url{fanruijia@stu.pku.edu.cn}, \url{guoquanwang@stu.pku.edu.cn}}). B. He is with the National Key Lab of Autonomous Intelligent Unmanned Systems, Tongji University, China (e-mail: \url{Hebin@tongji.edu.cn}).}
\thanks{This work is supported by the National Natural Science Foundation of China (No.62073004), Natural Science Foundation of Shenzhen (No.JCYJ20230807120801002). \newline}}



\maketitle

\begin{abstract}
Cloth-Changing Person Re-Identification (\ccreid) aims to accurately identify the target person in more realistic surveillance scenarios, where pedestrians usually change their clothing.
Despite great progress, limited cloth-changing training samples in existing \ccreid datasets still prevent the model from adequately learning cloth-irrelevant features.
In addition, due to the absence of explicit supervision to keep the model constantly focused on cloth-irrelevant areas, existing methods are still hampered by the disruption of clothing variations.
To solve the above issues, we propose an Identity-aware Dual-constraint Network (\modelnameshort) for the \ccreid task.
Specifically, to help the model extract cloth-irrelevant clues, we propose a Clothes Diversity Augmentation (CDA), which generates more realistic cloth-changing samples by enriching the clothing color while preserving the texture.
In addition, a Multi-scale Constraint Block (MCB) is designed, which extracts fine-grained identity-related features and effectively transfers cloth-irrelevant knowledge.
Moreover, a Counterfactual-guided Attention Module (CAM) is presented, which learns cloth-irrelevant features from channel and space dimensions and utilizes the counterfactual intervention for supervising the attention map to highlight identity-related regions.
Finally, a Semantic Alignment Constraint (SAC) is designed to facilitate high-level semantic feature interaction.
Comprehensive experiments on four \ccreid datasets indicate that our method outperforms prior state-of-the-art approaches.
\end{abstract}


\section{Introduction} \label{sec:1}
\IEEEPARstart{P}{erson} Re-Identification (\reid) focuses on identifying the same pedestrian across non-overlapping cameras, which has important application value in intelligent surveillance, smart shopping and human-computer interaction \cite{hou2021feature,si2022spatial,yeDeepLearningPerson2022,si2022hybrid,si2023tri}.
In the past few years, general person \reid methods \cite{wei2018glad,luo2019strong,zhao2020deep,zeng2020illumination} primarily concentrate on short-term scenarios, where the clothing and accessories of each pedestrian usually stay unchanged.
Therefore, most approaches achieve great performance by learning cloth-relevant features.
In practical application scenarios, surveillance videos of pedestrians are usually captured over a long time interval, where the clothing of pedestrians may change drastically.
This leads to the substantial performance reduction of traditional person \reid methods.

To mitigate the confusion of pedestrian identity caused by clothing variations, Cloth-Changing Person Re-Identification (\ccreid) has received extensive attention and continuous exploration. 
Existing \ccreid works mainly focus on feature disentanglement techniques or cloth-irrelevant feature learning with the attention mechanism.
For the former, cloth-relevant and cloth-irrelevant features are decoupled from pedestrian images \cite{eom2021gan, hong2021fine, cui2023dcr}.
For example, Hong \etal \cite{hong2021fine} utilize the pre-trained human parsing model to decouple fine-grained body shape features from original pedestrian images.
Cui \etal \cite{cui2023dcr} disentangle the cloth-relevant and cloth-irrelevant features by reconstructing human body parts.
However, learning to disentangle features is usually a time-consuming process.
For the latter, attention mechanisms are often exploited to extract cloth-agnostic features \cite{yang2023win,zhang2023multi,guo2023semantic,wang2022co}.
For example, Yang \etal \cite{yang2023win} propose a competitive attention strategy that gradually gathers detailed identity clues by employing multi-scale feature extraction across global, channel, and pixel levels.
Zhang \etal \cite{zhang2023multi} present a multi-branch structure and a differential feature attention module to extract discriminative biological features.
However, these methods often fail to learn fine-grained identity-relevant information as they do not consider imposing explicit constraints on the attention module.

Moreover, due to the difficulty of collecting and annotating cloth-changing samples, the clothing diversity of existing cloth-changing datasets is usually limited, which makes existing methods hardly extract cloth-irrelevant features and generalize well to real-life scenarios.
To solve this problem, some existing works improve clothing diversity from the data augmentation perspective, which can be categorized as image-based and feature-based augmentations \cite{shu2021semantic,jia2022complementary,zhao2023joint,han2023clothing}.
Shu \etal \cite{shu2021semantic} design a semantic-guided pixel sampling approach that randomly swaps clothing pixels among individuals to generate various cloth-changing samples.
However, the clothing colors and textures generated by this method are messy, which introduces much noise into pedestrian images and fails to simulate the actual scene.
Han \etal \cite{han2023clothing} present a clothing-change covariance estimation method and an identity-correlated augmentation strategy to implicitly augment clothing-change data in the feature space.
Nonetheless, it requires gathering clothing labels, which is cumbersome in practice.

\begin{figure}[t]
  \centering
  \includegraphics[width=\linewidth]{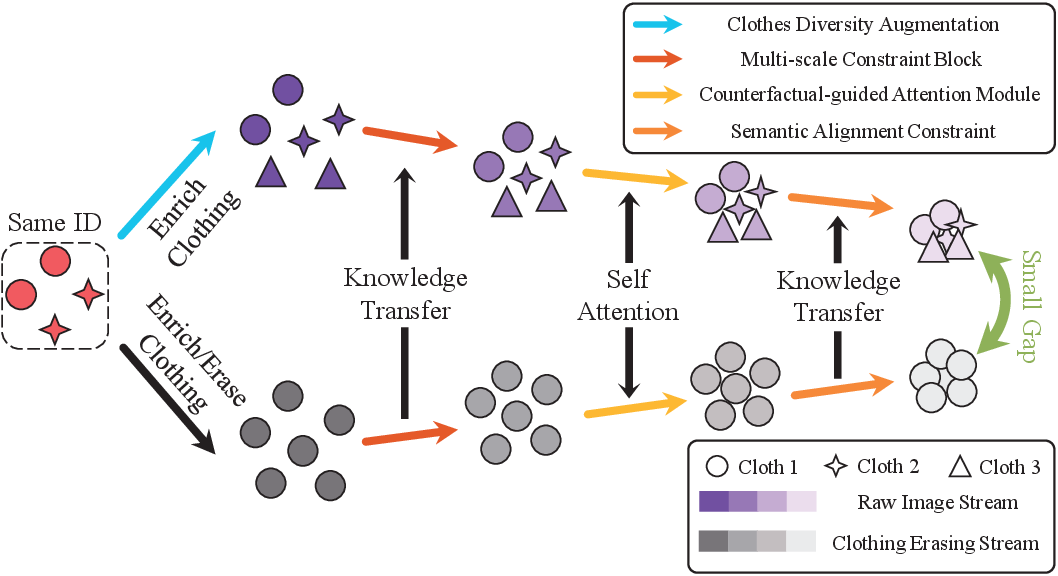}
  \caption{A schematic diagram of our approach. Data augmentation can effectively improve the variety of clothing and generate more realistic cloth-changing samples. The attention module can motivate the network to concentrate on areas not related to clothing. Dual constraints can realize cloth-agnostic knowledge transfer at both shallow and deep layers of the model.}
\label{fig:overall}
\end{figure}

To overcome the above weaknesses, we propose an Identity-aware Dual-constraint Network (\modelnameshort) for the \ccreid task.
Fig. \ref{fig:overall} is a schematic illustration of the proposed \modelnameshort.
There are two basic ideas of our approach: (1) Generating more realistic cloth-changing samples to enrich the clothing diversity of the training set, helping the model extract identity-related clues.
(2) Extracting discriminative cloth-irrelevant features by introducing dual-level constraints and effective attention modules, reducing the interference of clothing changes on the model.
Specifically, our \modelnameshort comprises a raw image stream and a clothing erasing stream.
In the raw image stream, the model primarily captures information about appearance, while in the clothing erasing stream, it focuses on learning cloth-agnostic identity features.
Firstly, we design a novel Clothes Diversity Augmentation (CDA), which does not require the collection of clothing labels and can generate more realistic cloth-changing images by preserving the texture of clothes while enriching the color.
Secondly, a Multi-scale Constraint Block (MCB) is proposed, which extracts features at global and local scales and utilizes the proposed hierarchical matching loss to effectively transfer the cloth-agnostic knowledge from the clothing erasing stream to the raw image stream at shallow layers of the model.
Thirdly, we introduce a Counterfactual-guided Attention Module (CAM) to mine cloth-agnostic clues from channel and spatial dimensions.
To supervise the model consistently focusing on cloth-irrelevant regions, counterfactual intervention and the proposed cloth-agnostic contrastive loss are employed to directly optimize the output features of the attention module.
Finally, a Semantic Alignment Constraint (SAC) is designed, which uses activation maps and saliency maps as mediums to achieve dual-stream feature alignment, facilitating cloth-irrelevant knowledge transfer at the high semantic level. 
During the inference stage, our \modelnameshort achieves efficient computation by removing the clothing erasing stream and utilizing features extracted from the raw image stream. \par
The main contributions are summarized as follows:
\begin{itemize}
\item For the cloth-changing person re-identification task, we present an identity-aware dual-constraint network named \modelnameshort. By imposing constraints at shallow and deep layers of the model, fine-grained cloth-irrelevant knowledge can be effectively transferred to learn robust identity-related features in an end-to-end manner.
\item A clothes diversity augmentation is proposed to generate realistic cloth-changing samples with preserving clothing textures, helping the model learn cloth-irrelevant features.
\item To facilitate cloth-agnostic knowledge transfer, a multi-scale constraint block and a semantic alignment constraint are designed, which perform mutual learning of appearance features and cloth-irrelevant features at low image level and high semantic level, respectively.
\item A counterfactual-guided attention module is introduced, which leverages counterfactual intervention and the proposed cloth-agnostic contrastive loss to keep the model consistently focused on clothing-independent regions. To our knowledge, this is the first work to apply counterfactual intervention to the \ccreid task.
\end{itemize}

\section{Related Work} \label{sec:2}
\subsection{Cloth-Changing Person Re-ID}
In practical scenarios, pedestrians are likely to change their outfits over a long period.
Therefore, \ccreid is proposed to solve the task of pedestrian identification in cloth-changing scenarios.
In recent years, some works leverage auxiliary biological features related to identity to tackle the challenge of \ccreid \cite{chen2021deep,jin2022cloth,wang2022co}.
Jin \etal \cite{jin2022cloth} extract gait information from the gait stream as auxiliary clues and discard it during the inference stage for efficient computation.
Wang \etal \cite{wang2022co} learn fine-grained body shape information and transfer it to the appearance branch by employing the cross-attention mechanism.
Recently, there have also been some works dedicated to decoupling cloth-related and cloth-unrelated features from pedestrian images to discard the impact of clothing changes \cite{yu2021apparel,xu2021adversarial,cui2023dcr,yang2023good}.
For example, Xu \etal \cite{xu2021adversarial} introduce an adversarial feature decoupling framework that incorporates intra-class reconstruction and inter-class adversary to separate cloth-relevant and cloth-irrelevant features.
Cui \etal \cite{cui2023dcr} present a component reconstruction disentanglement module that decouples cloth-related and cloth-unrelated features by reconstructing human body regions.
To mitigate clothing bias for robust \ccreid, yang \etal \cite{yang2023good} propose a causality-based auto-intervention model that removes cloth-relevant knowledge by subtracting features of the clothing branch.

Although existing works improve the performance of \ccreid to some extent, insufficient cloth-changing samples in public datasets still hinder these methods from adequately learning features that can effectively resist clothing changes.
Recently, several approaches propose data augmentation to increase clothing diversity in the training set \cite{shu2021semantic,han2023clothing,zhao2023joint,jia2022complementary}.
Zhao \etal \cite{zhao2023joint} view clothing changes as shifts in the style of images and utilize Instance Normalization (IN) to exchange feature statistics of samples within each identity at the instance level, synthesizing diverse and novel cloth-changing samples.
However, IN also decreases the variance between classes, which results in reduced discriminability of features.
Jia \etal \cite{jia2022complementary} simulate diverse appearance changes by randomly exchanging patches of clothing areas on pedestrians wearing different clothes, enhancing the robustness of the model against clothing variations.
Nevertheless, this method would destroy the texture of the clothing, and the clothing variations are limited to local patch areas.
Differently, our \modelnameshort generates rich cloth-changing samples without destroying the texture and preserves the discriminability of features.

\begin{figure*}[t]
  \centering
  \includegraphics[width=\linewidth]{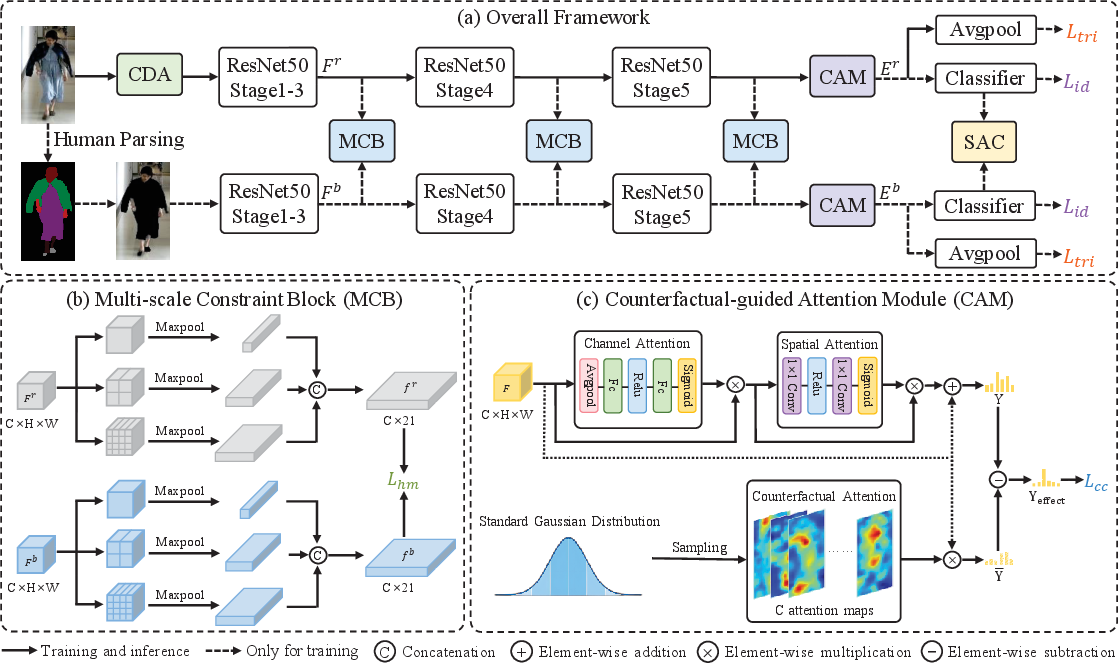}
  \caption{(a) Overall framework of our proposed identity-aware dual-constraint network (\modelnameshort). The top is the raw image stream, while the bottom is the clothing erasing stream. After performing Clothes Diversity Augmentation (CDA) on the training set, the augmented images and black-clothing images are input into backbone networks with the shared-weight architecture. Between stages of the backbone network, the proposed Multi-scale Constraint Blocks (MCBs) are inserted to extract multi-scale features and transfer cloth-independent knowledge to the raw image stream. After the backbone network, the Counterfactual-guided Attention Module (CAM) highlights identity-related features and utilizes counterfactual intervention to optimize the quality of learned attention maps. Finally, the Semantic Alignment Constraint (SAC) facilitates the raw image stream to learn cloth-agnostic semantic features. Moreover, each stream is supervised by triplet loss $L_{tri}$ and identity loss $L_{id}$. (b) Illustration of our proposed MCB. (c) Illustration of our proposed CAM.}
\label{fig:model}
\end{figure*}

\subsection{Counterfactual Analysis}
Counterfactual analysis is a part of causal theory \cite{yao2021survey}, which is often used to analyze the possible consequence of a decision or action by presenting a hypothetical situation.
Since causality can facilitate feature learning and enhance model interpretability, an increasing number of researchers have applied it to various fields, such as visual question answering \cite{chen2020counterfactual,niu2021counterfactual}, gait recognition \cite{dou2023gaitgci}, image recognition \cite{lopez2017discovering} and person re-identification \cite{li2022counterfactual,sun2023counterfactual}.
Rao \etal \cite{rao2021counterfactual} address fine-grained image recognition by employing counterfactual training to solve the bias of the spatial attention from the dataset.
Fu \etal \cite{fu2020sscr} incorporate counterfactual analysis into the iterative language-based image editing task, which adds counterfactual instructions to tackle the data scarcity issue.
In \cite{niu2021counterfactual}, the authors mitigate language bias in visual question answering by subtracting the direct impact of language from the overall causal impact. 
Goyal \etal \cite{goyal2019counterfactual} explore the attention areas of the depth vision system in image recognition tasks by generating counterfactual visual explanations.

In the field of person \reid, Li \etal \cite{li2022counterfactual} concentrate on emphasizing the affinity of the feature migration module through counterfactual intervention to improve the sub-optimal topology structure caused by the inherent limitations of the graph-based Visible-Infrared Re-ID (VI-ReID) model.
Sun \etal \cite{sun2023counterfactual} present a counterfactual attention alignment strategy to solve the VI-ReID task, which learns intra-modality information and aligns cross-modality features by using counterfactual analysis.
Different from existing works, in order to address the \ccreid task, our method utilizes counterfactual intervention to supervise cloth-irrelevant attention learning, helping the model extract identity-related discriminative features.

\section{Proposed Method} \label{sec:3}
\subsection{Overall Framework}
As mentioned above, decoupling cloth-related and cloth-unrelated features from original pedestrian images is time-consuming, and insufficient cloth-changing samples often hinder the model from learning discriminative cloth-irrelevant features.
Moreover, the original pedestrian images contain abundant cloth-irrelevant information that is underutilized by current methods.
To better learn identity-related clues from pedestrian images, we propose an Identity-aware Dual-constraint Network (\modelnameshort).
As shown in Fig. \ref{fig:model}(a), the \modelnameshort comprises two distinct streams: raw image stream and clothing erasing stream.
To increase the clothing diversity of pedestrian images, Clothes Diversity Augmentation (CDA) is performed on the training set, which enriches the color of clothes while preserving the texture.
Considering that clothing variations can disrupt the \ccreid model, we localize the clothing area by using the pre-trained human parsing model SCHPNet~\cite{li2020self} and specify pixels of the clothing region to a constant value (we adopt 0 for simplicity) for obtaining images where pedestrians dress black outfits.
The augmented pedestrian images are fed into the raw image stream to extract appearance information, while black-clothing images are input into the clothing erasing stream to learn cloth-irrelevant features.
ResNet50 \cite{he2016deep} initially pre-trained on the ImageNet dataset \cite{krizhevsky2017imagenet} is selected as our backbone network considering its relatively concise structure and superior performance, and backbone networks of two streams are set to have shared weights.
In addition, since the backbone is a crucial component for the model to extract identity-related features, the proposed Multi-scale Constraint Blocks (MCBs) are inserted between several stages of ResNet50 to learn multi-scale features and transfer the cloth-irrelevant knowledge from the clothing erasing stream to the raw image stream.
After the backbone network, the Counterfactual-guided Attention Module (CAM) extracts cloth-irrelevant features from channel and spatial dimensions and utilizes the counterfactual intervention to simultaneously enhance compactness within categories and discrepancy across categories in output features.
Subsequently, fine-grained appearance features and cloth-agnostic features are constrained to be consistent through the designed Semantic Alignment Constraint (SAC), facilitating the raw image stream to learn high-level cloth-irrelevant semantic features.
Finally, dual-stream features are fed into the classifier for identification.
The classifier comprises a batch normalization layer, a $1 \times 1$ convolution layer and global average pooling.
Besides, both streams are further optimized by the identity loss $L_{id}$ and triplet loss $L_{tri}$ to improve the classification ability of the model.
In conclusion, our approach can direct the network to focus on extracting features with increased resistance to clothing variations and adapt well to realistic long-term scenarios.

\begin{figure}[t]
  \centering
  \includegraphics[width=\linewidth,height=5.5cm]{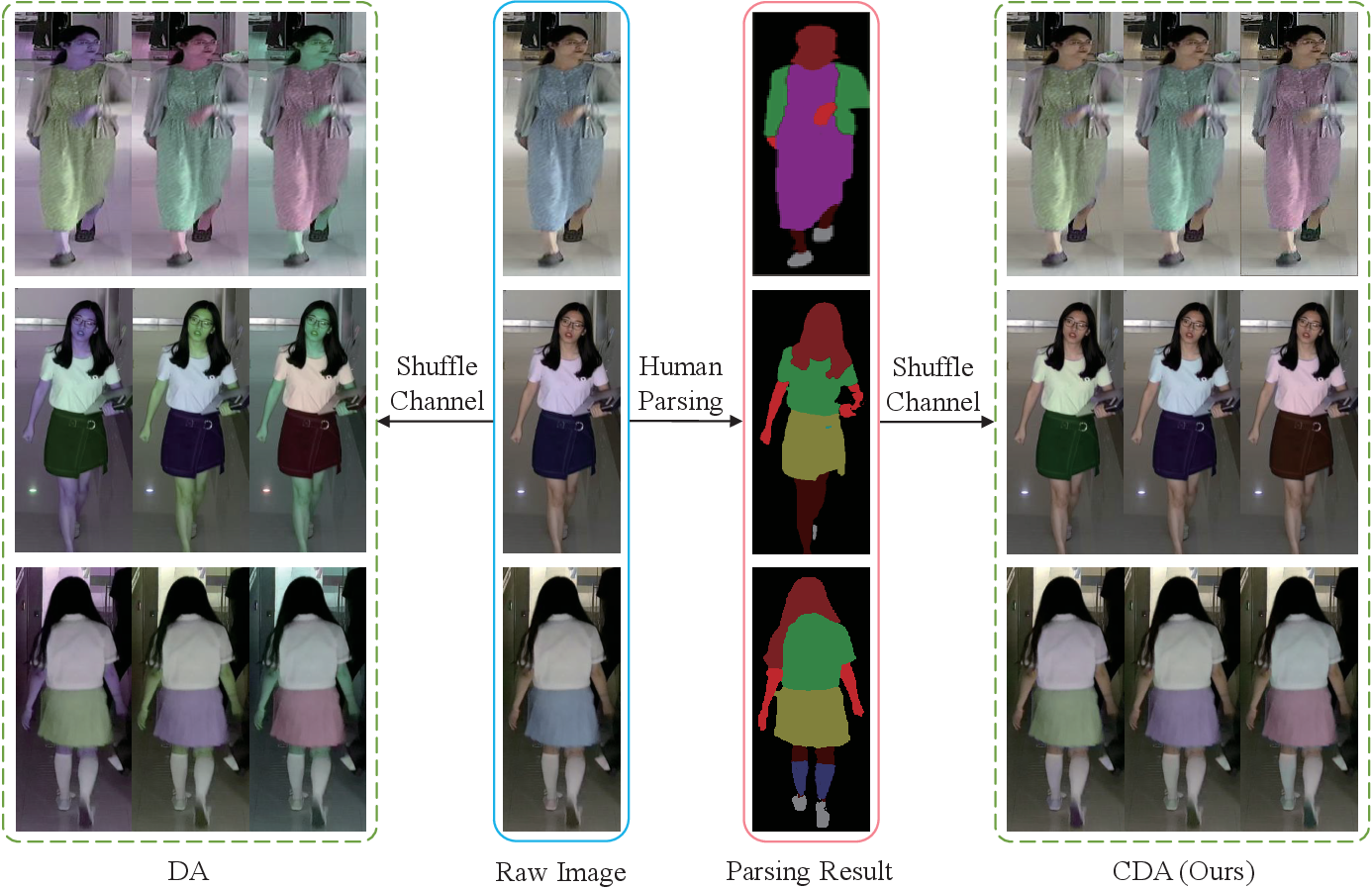}
  \caption{Comparison of the proposed clothes diversity augmentation (CDA) and domain augmentation (DA) \cite{chen2023unsupervised}. Each column represents a specific order of channels.}
\label{fig:augment}
\end{figure}

\subsection{Clothes Diversity Augmentation}
In realistic surveillance scenarios, the clothing changes of pedestrians are usually drastic.
However, the model trained on limited cloth-changing samples can not effectively learn cloth-agnostic information, resulting in performance degradation in real-life monitoring systems.
Therefore, we propose a Clothes Diversity Augmentation (CDA), aiming to enrich the clothing diversity of pedestrian images.
Specifically, a set of images in the training batch can be denoted as $\mathcal{X}=\left\{x_1^1, x_i^1, x_K^1, \cdots, x_1^p, x_i^p, x_K^p, \cdots, x_1^P, x_i^P, x_K^P\right\}$, comprising $P$ distinct identities with $K$ samples per identity.
Here, $x_i^p$ represents the $i$-th sample of the $p$-th identity.
For locating the clothing area in pedestrian images, we utilize the pre-trained SCHPNet~\cite{li2020self} to generate a human parsing result, and each pixel in the parsing result is a semantic label corresponding to cloth-related or cloth-unrelated regions.
After the pixel coordinates of the clothing area are obtained from the parsing result, we randomly shuffle and recompose RGB channels in the clothing region, as depicted in Fig. \ref{fig:augment}.
This data augmentation can be formulated as:
\begin{equation}
\resizebox{0.91\hsize}{!}{$\begin{gathered}
\tilde{x}_i^p=\left\{x_{(a, b),}^{c_1}, x_{(a, b)}^{c_2}, x_{(a, b)}^{c_3} \mid\left\{c_1, c_2, c_3\right\} \in \text{shuffle} \{R,G, B\} \right\} ,
\end{gathered}$}
\end{equation}
where $\tilde{x}_i^p$ indicates the augmented cloth-changing sample, and $(a, b)$ denotes the horizontal and vertical coordinates of pixels in the clothing region. 
The $\{c_1,c_2,c_3\}$ represents three channels of the image, and operation shuffle denotes randomly changing the order of channels, which contains all possibilities including the original $\{R, G, B\}$ order.
Different from Domain Augmentation (DA) \cite{chen2023unsupervised} that randomly shuffles and reorganizes the RGB channel of pedestrian images, our proposed CDA only changes the order of RGB channels in cloth-related areas, avoiding the possible negative impact on the training of \ccreid model from changing the identity-related regions such as exposed head and skin.
Moreover, compared with semantic-guided pixel sampling \cite{shu2021semantic}, we enrich the color of clothing without destroying the texture, generating more realistic pedestrian images. 
In conclusion, the proposed CDA can significantly increase the variety of clothing variations, encouraging the model to pay more attention to cloth-unrelated regions during training.

\subsection{Multi-scale Constraint Block}
Pedestrian images can be regarded as containing fine-grained identity information that is crucial for identification. 
However, most existing networks lose a lot of discriminative detailed features after the backbone network.
Therefore, a Multi-scale Constraint Block (MCB) is presented to fully learn fine-grained identity-related features and transfer cloth-irrelevant knowledge to the raw image stream.
Specifically, as illustrated in Fig. \ref{fig:model}(a), after the dual-stream features $F^r, F^b \in \mathbb{R}^{C \times H \times W}$ pass through the stage 1-3 of ResNet50, they are taken as input to MCB, where $C$ denotes the number of channels, $H$ denotes height, and $W$ represents width.
As shown in Fig. \ref{fig:model}(b), we perform the max pooling operation on $F^r$ and $F^b$ at different spatial scales, dividing them into 1, 4 and 16 parts respectively to extract features of different granularities from coarse to fine.
Then, we reduce the last dimension of pooled features and concatenate them to obtain $f^r, f^b \in \mathbb{R}^{C \times 21}$.
After obtaining the multi-scale features, a new hierarchical matching loss $L_{hm}$ is proposed to supervise the alignment of features from two streams at different layers of the backbone, which can be defined as:
\begin{equation}
\resizebox{0.53\hsize}{!}{$\begin{gathered}
L_{hm}=\frac{1}{N} \sum_{n=1}^N \sum_{m=1}^M\left(f^r_m-f^b_m\right)^2 ,
\end{gathered}$}
\end{equation}
where $N$ denotes the batch size, $M$ denotes the number of MCB.
Besides, $f^r_m$ and $f^b_m$ represent the output features of the $m$-th MCB from the raw image stream and clothing erasing stream, respectively.
Under the supervision of $L_{hm}$, dual-stream features can be effectively pulled into the public space, enhancing the ability of the model to resist clothing changes.
Notably, the output features of MCB $f^r_m$ and $f^b_m$ are not used as inputs to the next stage of backbone networks, but are only used as auxiliary features to achieve knowledge transfer, \ie, MCB does not affect the forward process of backbones.

\subsection{Counterfactual-guided Attention Module}
In order to fully extract cloth-agnostic information, a Counterfactual-guided Attention Module (CAM) is designed, which learns cloth-irrelevant features from channel and spatial dimensions respectively, as shown in Fig. \ref{fig:model}(c).
Specifically, motivated by \cite{hu2018squeeze}, the channel attention part includes a global average pooling layer and two fully connected layers, and each fully connected layer is followed by relu and sigmoid activation functions, respectively.
The purpose of this design is to effectively condense global spatial information into a channel descriptor, ensuring a detailed capture of dependencies at the channel level.
To reduce complexity and improve generalization ability, the two fully connected layers reduce the number of channels to 1/8 of the original and revert to the previous number, respectively.
The spatial attention part comprises two $1 \times 1$ convolution layers, each followed by relu and sigmoid activation functions separately.
It generates a spatial descriptor by compressing the number of channels of the feature to 1 and efficiently mining the regions associated with the identity in the spatial dimension.
Then, channel and spatial descriptors are element-wise multiplied with input features to highlight identity-related information.
Finally, the output of the spatial attention part and the original input feature $F$ are residually connected to achieve the fusion of multi-level features.
The above process can be formulated as:
\begin{equation}
\resizebox{0.76\hsize}{!}{$\begin{gathered}
Y=F+SA\left(F * CA\left(F\right)\right) *\left(F * CA\left(F\right)\right) ,
\end{gathered}$}
\end{equation}
where $CA$ denotes the channel attention, and $SA$ represents the spatial attention.

Intuitively, the effectiveness of the learned features depends on the performance of the attention module.
Traditional likelihood methods often directly optimize prediction results and ignore the impact of attention learning on output features, lacking effective supervision of the attention module.
Causal inference \cite{yao2021survey} provides a new perspective: imposing explicit constraints on the attention module by analyzing causal relationships between variables.
As a result, we bring in counterfactual intervention in causal inference to assess the validity of learned attention and optimize the model by motivating it to generate more effective attention maps.
Specifically, as shown in Fig. \ref{fig:casual}, $F$ denotes the input feature, $A$ denotes the learned attention map, and $Y$ denotes the output.
The path $F \rightarrow A$ denotes that $F$ can affect the value of $A$, indicating a causal relationship between them. 
Similarly, other paths also denote the same causal connection.
With the introduction of the causal graph, we are able to explore causal relationships through the direct modification of multiple variable values.
In the literature of causal inference, this action is formally known as intervention, denoted as $do(\cdot)$.
Here, $do(A=\overline{A})$ means replacing the $A$ with $\overline{A}$ to eliminate the causal relationship between the input feature $F$ and the attention map $A$.
Since attention $\overline{A}$ is not the real attention map learned by the model, $do(A=\overline{A})$ is called counterfactual intervention.
Following \cite{tang2020unbiased,li2022counterfactual}, the practical influence of learned attention map on the output $Y_{\text {effect }}$ can be described as the discrepancy between the actual output $Y$ and its counterfactual alternative $\overline{Y}$:

\begin{figure}[t]
  \centering
  \includegraphics[width=\linewidth]{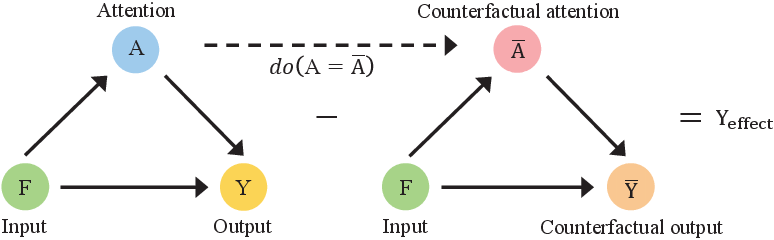}
  \caption{Elaboration of causal relationship and counterfactual intervention in the causal graph.}
\label{fig:casual}
\end{figure}

\begin{equation}
\resizebox{0.5\hsize}{!}{$\begin{gathered}
Y_{\text{effect}}=Y_{F,A}-\overline{Y}_{F,\operatorname{do}(A=\overline{A})} ,
\end{gathered}$}
\end{equation}
where $Y_{F,A}$ denotes the true output calculated from the input $F$ and the learned attention map $A$, and $\overline{Y}_{F,\operatorname{do}(A=\overline{A})}$ represents the output computed by $F$ and counterfactual attention $\overline{A}$.
In practice, we can choose reversed attention, uniform attention and random attention as counterfactual attention \cite{rao2021counterfactual}.
For simplifying the computation, we sample from the standard Gaussian distribution to obtain counterfactual attention $\overline{A}$, $\ie$, $\overline{A} \sim \mathcal{N}(0, 1)$.
Intuitively, we can measure the performance of attention by assessing how correct attention enhances the discriminative power of output features when compared to noisy attention.
Gaussian distributed attention has much noise and can not effectively highlight identity-related information.
Therefore, $Y_{\text{effect}}$ can be utilized as a measure to assess the quality of learned attention.\par
To supervise the learning of the attention module, a novel cloth-agnostic contrastive loss $L_{cc}$ is proposed.
For each anchor within a mini-batch, all images with the same identity are considered positive samples, and images with different identities from the anchor are negative samples.
We use cosine similarity as a metric of the affinity for each sample pair.
The $L_{cc}$ can be formulated as follows:
\begin{equation}
\resizebox{0.85\hsize}{!}{$\begin{gathered}
L_{cc} = -\frac{1}{N} \sum_{i=1}^N L_i\xspace, \\
L_i=\frac{1}{N_j} \sum_{j=1}^N \mathbbm{1}_{i \neq j} \cdot \mathbbm{1}_{y_i=y_j} \cdot \log \left(\exp \left(z_i \cdot z_j\right) \slash D \right)\xspace, \\
D = \exp \left(z_i \cdot z_j\right)+\sum_{k=1}^N \mathbbm{1}_{i \neq k} \cdot \mathbbm{1}_{y_i \neq y_k} \cdot \exp \left(z_i \cdot z_k\right)\xspace ,
\end{gathered}$}
\end{equation}
where $N$ denotes the batch size, and $N_j$ represents the number of positive samples for each anchor.
Here, $z$ denotes the feature after performing average pooling to $Y_{\text{effect}}$, and $y$ denotes the person identity label.
The symbol $\mathbbm{1}_{(\cdot)}$ represents an indicator function, which takes on a value of 1 when the condition $(\cdot)$ is satisfied and 0 otherwise.
Under the supervision of $L_{cc}$, features belonging to the same identity are pulled together, while features belonging to different identities are pushed away.
By optimizing the new objective, the attention module can extract features that are more robust to clothing changes and avoid sub-optimal results.

\subsection{Semantic Alignment Constraint}
\begin{figure}[t]
  \centering
  \includegraphics[width=\linewidth]{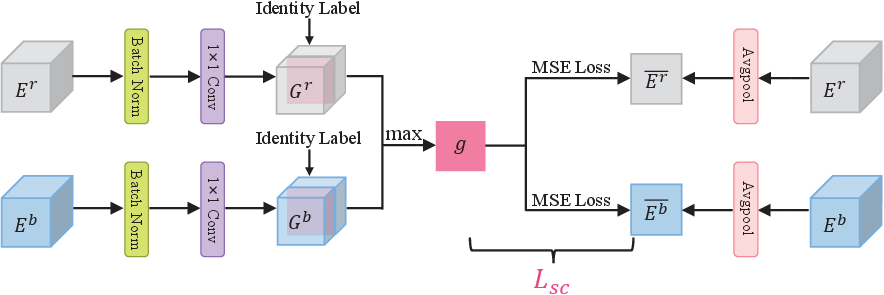}
  \caption{Illustration of the Semantic Alignment Constraint (SAC).}
\label{fig:sc}
\end{figure}

To achieve cloth-irrelevant knowledge transfer at the semantic level, a Semantic Alignment Constraint (SAC) is designed, as illustrated in Fig. \ref{fig:sc}.
Specifically, a batch normalization layer is utilized to process features from two streams $E^r$ and $E^b$.
Subsequently, these features are fed into a $1 \times 1$ convolution layer, resulting in the generation of class activation maps $G^r$ and $G^b \in \mathbb{R}^{I \times H \times W}$, where $I$ represents the count of pedestrian identities.
Given that the class activation map indicates what areas the model focuses on when distinguishing each category, we choose the feature map that corresponds to the true identity label in the channel dimension.
Then, we compare each pixel in the two feature maps and select the maximum value to obtain a more effective supervision signal $g \in \mathbb{R}^{H \times W}$.
At the same time, we conduct average pooling on the features $E^r$ and $E^b$ along the channel dimension to generate saliency maps $\overline{E^r}$ and $\overline{E^b} \in \mathbb{R}^{H \times W}$, which essentially indicate attention regions of the network.
To achieve semantic-level feature alignment, we extend the idea of \cite{guo2023semantic} and impose a semantic consistency loss $L_{sc}$, which can be denoted as:
\begin{equation} \label{eq:sc}
L_{s c}=\frac{1}{N} \sum_{n=1}^N[(g-\overline{E^r})^2+(g-\overline{E^b})^2] \xspace,
\end{equation}
where $N$ denotes the batch size.
By approximating the saliency maps, the $L_{sc}$ encourages the raw image stream to fully take advantage of discriminative features from the clothing erasing stream, enhancing the robustness under clothing variations.

\subsection{Training and Inference}
As depicted in Fig. \ref{fig:model}(a), we also employ triplet loss \cite{hermans2017defense} and identity loss \cite{sun2018beyond} in each stream to optimize the model.
Identity loss measures the performance of person \reid models from the perspective of image classification.
For the $i$-th image of the training set $x_i$ with the identity label $y_i$, the probability that $x_i$ belongs to class $y_i$ is predicted by a classifier and softmax function, denoted as $p\left(y_i \mid x_i\right)$.
Subsequently, the identity loss is calculated using cross-entropy:
\begin{equation}
L_{id}=-\frac{1}{N} \sum_{i=1}^N \log \left(p\left(y_i \mid x_i\right)\right) ,
\end{equation}
where $N$ represents the batch size.
Triplet loss evaluates the performance of person \reid models by analyzing the feature distribution of intra-class and inter-class samples.
Different from the proposed cloth-agnostic contrastive loss, the primary target of the triplet loss is that the distance between the farthest positive sample pair is at least one pre-defined margin smaller than the distance between the closest negative sample pair. 
The triplet loss with a margin parameter is denoted as:
\begin{equation}
L_{t r i}(i, j, k)=\frac{1}{N} \sum_{i=1}^N \max \left(\rho+d_{i j}-d_{i k}, 0\right) ,
\end{equation}
where $N$ denotes the batch size.
Here, $d_{ij}$ represents the Euclidean distance between the farthest positive sample pair, $d_{ik}$ indicates the Euclidean distance between the closest negative sample pair, and $\rho$ denotes the margin parameter.
Identity loss operates in the cosine space, aiming to optimize the classification capability of the model.
Differently, triplet loss operates in Euclidean space, focusing on refining features by enhancing both intra-class compactness and inter-class discrepancy.
The overall training loss of the framework is formulated as follows:
\begin{equation} \label{eq:total}
L=\lambda_1(L_{hm}+L_{sc})+\sum_{n \in Z}(\lambda_2L_{cc}^n+L_{id}^n+L_{tri}^n) \xspace,
\end{equation}
where $Z=\{r,b\}$ corresponds to the raw image stream and clothing erasing stream, respectively.
Hyper-parameters $\lambda_1$ and $\lambda_2$ are utilized to control the weight of loss functions during the training process.
The proposed model is trained end-to-end, eliminating the requirement for additional training procedures.
During the inference stage, the clothing erasing stream is discarded to effectively decrease dependence on the human parsing model and reduce computational costs.

\section{Experiments and Discussions} \label{sec:experiment}
\subsection{Datasets and Evaluation Metrics}
To evaluate our proposed method, extensive experiments are performed on four public \ccreid datasets: LTCC \cite{qian2020long}, PRCC \cite{yang2019person}, Vc-Clothes \cite{wan2020person}, and DeepChange \cite{xu2023deepchange}.
In addition, we conduct experiments on the Market-1501 dataset \cite{zheng2015scalable} to validate the effectiveness of the proposed method in cloth-consistent scenarios.
The LTCC dataset consists of 17,119 indoor pedestrian images of 152 identities, exhibiting 478 distinct outfits.
It is considered one of the most difficult \ccreid datasets due to multiple scenes taken by 12 cameras, drastic clothing changes (up to 14 outfits per person) and significant illumination variations.
The PRCC dataset comprises 33,698 indoor images captured by 3 cameras, including 221 pedestrians.
Each pedestrian with the same identity in the first two cameras shows the same clothing, while the pedestrian in the last camera wears different clothes.
The Vc-Clothes dataset is synthesized using a game engine called GTA5 and contains 19,060 images of 512 people.
This dataset is collected using 4 cameras, and each pedestrian wears up to 3 distinct outfits.
The DeepChange dataset is a large-scale collection of outdoor images captured by 17 surveillance cameras.
It includes 178,407 images from 1,121 identities.
The Market-1501 dataset contains 32,668 images of 1,501 persons collected from 6 camera views. 
No pedestrians change their clothes in Market-1501.

\begin{table*}[t]
\caption{Comparison with state-of-the-art methods on the cloth-changing datasets LTCC, PRCC and Vc-Clothes. 'Parsing', 'Pose', 'Gait', 'Contour' and 'Clothes ID' represent human parsing, human poses, gait information, body contour and clothes label, respectively. \textbf{Bold} numbers are the optimal results, same as the following.}
\begin{center}
\setlength{\tabcolsep}{0.3mm}
\renewcommand{\arraystretch}{1.1}
\begin{tabular}{c|c|c|c|cccc|cccc|cccc}
\hline
\multirow{3}{*}{Method} & \multirow{3}{*}{Venue} & \multirow{3}{*}{Size} & \multirow{3}{*}{Assistance} & \multicolumn{4}{c|}{LTCC} & \multicolumn{4}{c|}{PRCC} & \multicolumn{4}{c}{Vc-Clothes}\\ \cline{5-16} 
& & & & \multicolumn{2}{c|}{General} & \multicolumn{2}{c|}{Cloth-changing} & \multicolumn{2}{c|}{Same-clothes} & \multicolumn{2}{c|}{Cloth-changing} & \multicolumn{2}{c|}{General} & \multicolumn{2}{c}{Cloth-changing} \\ \cline{5-16}
& & & & Rank-1 & \multicolumn{1}{c|}{mAP} & Rank-1 & mAP & Rank-1 & \multicolumn{1}{c|}{mAP} & Rank-1 & mAP & Rank-1 & \multicolumn{1}{c|}{mAP} & Rank-1 & mAP \\ \hline
HACNN \cite{li2018harmonious} & CVPR'18 & $256\times128$ & None & 60.2 & \multicolumn{1}{c|}{26.7} & 21.6 & 9.3 & 82.5 & \multicolumn{1}{c|}{84.8} & 21.8 & 23.2 & 68.6 & \multicolumn{1}{c|}{69.7} & 49.6 & 50.1 \\
PCB \cite{sun2018beyond} & ECCV'18 & $384\times192$ & None & 65.1 & \multicolumn{1}{c|}{30.6} & 23.5 & 10.0 & 99.8 & \multicolumn{1}{c|}{97.0} & 41.8 & 38.7 & 87.7 & \multicolumn{1}{c|}{74.6} & 62.0 & 62.2 \\
IANet \cite{hou2019interaction} & CVPR'19 & $384\times192$ & None & 63.7 & \multicolumn{1}{c|}{31.0} & 25.0 & 12.6 & 99.4 & \multicolumn{1}{c|}{98.3} & 46.3 & 45.9 & - & \multicolumn{1}{c|}{-} & - & - \\
ISP \cite{zhu2020identity} & ECCV'20 & $256\times128$ & None & 66.3 & \multicolumn{1}{c|}{29.6} & 27.8 & 11.9 & 92.8 & \multicolumn{1}{c|}{-} & 36.6 & - & 94.5 & \multicolumn{1}{c|}{94.7} & 72.0 & 72.1 \\ 
FSAM \cite{hong2021fine} & CVPR'21 & $256\times128$ & Parsing+Pose & 73.2 & \multicolumn{1}{c|}{35.4} & 38.5 & 16.2 & - & \multicolumn{1}{c|}{-} & - & - & 94.7 & \multicolumn{1}{c|}{\textbf{94.8}} & 78.6 & 78.9 \\
CAL \cite{gu2022clothes} & CVPR'22 & $384\times192$ & Clothes ID & 74.2 & \multicolumn{1}{c|}{40.8} & 40.1 & 18.0 & 100 & \multicolumn{1}{c|}{99.8} & 55.2 & 55.8 & 92.9 & \multicolumn{1}{c|}{87.2} & 81.4 & 81.7 \\
GI-ReID \cite{jin2022cloth} & CVPR'22 & $256\times128$ & Parsing+Gait & 63.2 & \multicolumn{1}{c|}{29.4} & 23.7 & 10.4 & - & \multicolumn{1}{c|}{-} & - & - & - & \multicolumn{1}{c|}{-} & 64.5 & 57.8 \\
M2Net \cite{liu2022long} & MM'22 & $256\times128$ & Parsing+Contour+Clothes ID & - & \multicolumn{1}{c|}{-} & - & - & 99.5 & \multicolumn{1}{c|}{99.1} & 59.3 & 57.7 & - & \multicolumn{1}{c|}{-} & - & - \\
ACID \cite{yang2023win}& TIP'23 & $256\times128$ & None & 65.1 & \multicolumn{1}{c|}{30.6} & 29.1 & 14.5 & - & \multicolumn{1}{c|}{-} & - & - & 95.1 & \multicolumn{1}{c|}{94.7} & 84.3 & 74.2 \\
AIM \cite{yang2023good} & CVPR'23 & $384\times192$ & Clothes ID & 76.3 & \multicolumn{1}{c|}{41.1} & 40.6 & 19.1 & 100 & \multicolumn{1}{c|}{\textbf{99.9}} & 57.9 & 58.3 & - & \multicolumn{1}{c|}{-} & - & - \\
CCFA \cite{han2023clothing}& CVPR'23 & $384\times128$ & Clothes ID & 75.8 & \multicolumn{1}{c|}{42.5} & 45.3 & 22.1 & 99.6 & \multicolumn{1}{c|}{98.7} & 61.2 & 58.4 & - & \multicolumn{1}{c|}{-} & - & - \\
IMS-GEP \cite{zhao2023joint}& TMM'23 & $384\times192$ & None & - & \multicolumn{1}{c|}{-} & 43.4 & 18.2 & 99.7 & \multicolumn{1}{c|}{99.8} & 57.3 & \textbf{65.8} & - & \multicolumn{1}{c|}{-} & 81.8 & 81.7 \\
AFL \cite{liu2023clothes}& TMM'23 & $384\times192$ & None & 74.4 & \multicolumn{1}{c|}{39.1} & 42.1 & 18.4 & 100 & \multicolumn{1}{c|}{99.7} & 57.4 & 56.5 & 93.9 & \multicolumn{1}{c|}{88.3} & 82.5 & 83.0 \\ 
DCR-ReID \cite{cui2023dcr}& TCSVT'23 & $384\times192$ & Parsing+Contour+Clothes ID & 76.1 & \multicolumn{1}{c|}{42.3} & 41.1 & 20.4 & 100 & \multicolumn{1}{c|}{99.7} & 57.2 & 57.4 & - & \multicolumn{1}{c|}{-} & - & - \\ \hline
\modelnameshort(Ours) & - & $256\times128$ & Parsing & 76.8 & \multicolumn{1}{c|}{48.3} & 51.6 & 33.5 & 100 & \multicolumn{1}{c|}{98.8} & 62.7 & 62.5 & 95.2 & \multicolumn{1}{c|}{90.7} & 91.3 & 86.8 \\
\modelnameshort(Ours) & - & $384\times192$ & Parsing & \textbf{77.1} & \multicolumn{1}{c|}{\textbf{49.9}} & \textbf{53.1} & \textbf{35.9} & \textbf{100} & \multicolumn{1}{c|}{98.8} & \textbf{64.9} & 64.6 & \textbf{96.0} & \multicolumn{1}{c|}{91.6} & \textbf{92.5} & \textbf{87.9} \\ \hline
\end{tabular}
\label{tab:sota}
\end{center}
\end{table*}
The used evaluation metrics include Rank-1 accuracy and mean Average Precision (mAP), and all experiments are performed under the multi-shot evaluation protocol.
Following \cite{gu2022clothes}, test settings are classified as (1) \textbf{same-clothes setting:} only cloth-consistent ground truth samples are used to calculate accuracy, (2) \textbf{cloth-changing setting:} only cloth-changing ground truth samples are used to calculate accuracy, and (3) \textbf{general setting:} both cloth-changing and cloth-consistent ground truth samples are used to calculate accuracy.
Following \cite{liu2023clothes}, we report Rank-1 accuracy and mAP on the LTCC and Vc-Clothes datasets under the general and cloth-changing settings.
At the same time, we report Rank-1 accuracy and mAP on the PRCC dataset under the same-clothes and cloth-changing settings \cite{gu2022clothes, yang2019person}.
For DeepChange, Rank-1 accuracy and mAP under the general setting are reported \cite{gu2022clothes}.
Since there is no cloth-changing sample in Market-1501, Rank-1 accuracy and mAP are reported following the standard evaluation protocol \cite{sun2018beyond,zhao2023joint}.

\subsection{Implementation Details}
Following the general routine \cite{luo2019bag}, we cancel the downsampling of the last stage in the backbone network to mitigate the loss of effective features in the spatial dimension and remove the final pooling layer and fully connected layer.
The backbone networks of two streams share weights.
The resolution of input images is set to $384\times192$ following \cite{qian2020long,guo2023semantic}. 
Besides the proposed CDA, the used data augmentation methods also include random erasing \cite{zhong2020random}, random cropping and random horizontal flipping. 
The batch size of training is 128, including 16 persons and 8 images for each person.
The training of the model employs the AdamW optimizer \cite{loshchilov2018fixing} across a span of 150 epochs.
For the learning rate, we adopt the warm-up strategy \cite{luo2019bag}, $\ie$, it starts at $3.5\times 10^{-6}$ and linearly increases to $3.5\times 10^{-4}$ over the first 10 epochs, then it is divided by 10 at the 40-th and 80-th epochs, respectively.
In Eq. \eqref{eq:total}, $\lambda_1$ is set to 0.01 on the LTCC, Vc-Clothes, DeepChange and Market-1501 datasets and to 0.05 on the PRCC dataset. 
$\lambda_2$ is set to 0.1 on the LTCC, PRCC, Vc-Clothes and Market-1501 datasets and to 0.01 on the Deepchange dataset.
The triplet loss margin $\alpha$ is determined to be 0.3 following \cite{han2023clothing,guo2023semantic}.

\subsection{Comparison with State-of-the-art Methods}
We evaluate our proposed \modelnameshort on the LTCC, PRCC, Vc-Clothes and DeepChange datasets for comparison with recent state-of-the-art methods, and results are shown in Table \ref{tab:sota}.
Following \cite{yang2023good}, we also perform experiments with an input size of $256\times128$ for a fairer comparison.
It can be observed that the majority of \ccreid methods (from FSAM \cite{hong2021fine} to DCR-ReID \cite{cui2023dcr}) outperform general \reid methods (from HACNN \cite{li2018harmonious} to ISP \cite{zhu2020identity}) attributed to the addition of auxiliary modules or loss functions that help the model learn cloth-irrelevant features.
Compared with these approaches, the proposed \modelnameshort achieves superior performance under the cloth-change setting.
For example, on the LTCC dataset, the proposed method is superior to sub-optimal CCFA \cite{han2023clothing} by 7.8\% and 13.8\% on Rank-1 and mAP, respectively.
On the PRCC dataset, our approach outperforms CCFA by 3.7\% Rank-1.
For mAP, our method also realizes a result comparable to that of SOTA methods.
Likewise, on the Vc-Clothes dataset, our method also surpasses ACID \cite{yang2023win} by 7.0\% Rank-1 and AFL \cite{liu2023clothes} by 4.9\% mAP, respectively.
While primarily designed for the \ccreid task, the proposed method also achieves competitive performance with SOTA methods under the general and same-clothes settings.
\begin{table}[t]
\caption{Comparison with state-of-the-art methods on DeepChange. It is performed under the general setting.}
\vspace{-1.8mm}
\begin{center}
\setlength{\tabcolsep}{3mm}
\renewcommand{\arraystretch}{1.1}
\begin{tabular}{c|c|c|cc}
\hline
\multirow{2}{*}{Method} & \multirow{2}{*}{Venue} & \multirow{2}{*}{Size} & \multicolumn{2}{c}{DeepChange} \\ \cline{4-5} 
& & & Rank-1 & mAP \\ \hline
MGN \cite{wang2018learning} & MM'18 & $256\times128$ & 25.4 & 9.8 \\
ABD-Net \cite{chen2019abd} & ICCV'19 & $256\times128$ & 24.2 & 8.5 \\
OSNet \cite{zhou2019omni} & ICCV'19 & $256\times128$ & 39.7 & 10.3 \\
RGA \cite{zhang2020relation} & CVPR'20 & $256\times128$ & 28.9 & 8.6 \\
ReIDCaps \cite{huang2019beyond} & TCSVT'20 & $384\times192$ & 39.5 & 11.3 \\
Trans-reID \cite{he2021transreid} & ICCV'21 & $256\times128$ & 35.9 & 14.4 \\
CAL \cite{gu2022clothes} & CVPR'22 & $384\times192$ & 54.0 & 19.0 \\ 
SCNet \cite{guo2023semantic} & MM'23 & $384\times192$ & 53.5 & 18.7 \\ \hline
Ours & - & $256\times128$ & 55.3 & 21.7 \\
Ours & - & $384\times192$ & \textbf{58.2} & \textbf{24.5} \\ \hline
\end{tabular}
\label{tab:deepchange}
\end{center}
\end{table}
It is worth mentioning that some state-of-the-art methods utilize both the pre-trained human parsing model and contour/pose extractor to avoid interference from clothing changes, and the computational cost of these approaches is significantly higher than our proposed method.
Some methods also heavily rely on the collection of clothes labels, which is often time-consuming in practice.
Our proposed \modelnameshort can achieve better performance without collecting clothes labels.
In addition, we also conduct a comparison of the proposed \modelnameshort with the state-of-the-art methods on the large-scale DeepChange dataset, and the results are illustrated in Table \ref{tab:deepchange}.
It can be observed that our method is significantly superior to other recent SOTA methods under the general setting.
Compared with CAL \cite{gu2022clothes}, the proposed \modelnameshort outperforms it by 4.2\% Rank-1 and 5.5\% mAP, respectively.
The above experimental results show that the proposed method can effectively extract fine-grained identity-related features and is more robust to clothing changes.
Moreover, the additional experiment on Market-1501 is conducted to validate the suitability of our method in cloth-consistent scenarios, as shown in Table \ref{tab:market}.
According to the result, the proposed \modelnameshort is superior to other state-of-the-art general person \reid methods PCB \cite{sun2018beyond}, OSNet \cite{zhou2019omni}, ISP and CDNet \cite{li2021combined}.
It also surpasses the state-of-the-art \ccreid methods.
Notably, in the inference stage, our \modelnameshort removes the clothing erasing stream and only utilizes features from the raw image stream to improve computational efficiency.
Therefore, the proposed method can effectively solve the \ccreid task and is suitable for deployment in real-world surveillance applications.

\subsection{Ablation Study} \label{sec:ablation}
Extensive ablation studies are conducted to explore the impact of each proposed component.
Firstly, a new baseline is established.
Then, we conduct ablation experiments on the LTCC and PRCC datasets, respectively.
Finally, we study the influence of varying hyper-parameters.
Notably, all ablation studies are performed under the cloth-changing setting except for the hyper-parameter experiment.
\subsubsection{Baseline}
We introduce a new baseline as a basic reference for the ablation study.
Specifically, the baseline is a dual-stream architecture consisting of the raw image stream and the clothing erasing stream, taking original pedestrian images and black-clothing images as inputs, respectively.
In each stream, we adopt the ResNet50 \cite{he2016deep} pre-trained on the ImageNet dataset \cite{krizhevsky2017imagenet} as the fundamental backbone network.
The two backbone networks share weights.
After input images pass through the backbone networks of two streams, the features are directly fed into classifiers.
Each stream is jointly optimized using only the identity loss and triplet loss.
Data augmentation is achieved through random erasing, random horizontal flipping and random cropping.
Other experimental settings are consistent with Section \ref{sec:experiment}.B.
Therefore, the baseline is essentially a dual-stream network, which can alleviate the interference caused by clothing variations to some extent.

\begin{table}[t]
\caption{Comparison with state-of-the-art methods on the cloth-consistent dataset Market-1501.}
\vspace{-1.8mm}
\begin{center}
\setlength{\tabcolsep}{3.3mm}
\renewcommand{\arraystretch}{1.1}
\begin{tabular}{c|c|c|cc}
\hline
\multirow{2}{*}{Method} & \multirow{2}{*}{Venue} & \multirow{2}{*}{Size} &\multicolumn{2}{c}{Market-1501} \\ \cline{4-5} 
& & & Rank-1 & mAP \\ \hline
PCB \cite{sun2018beyond} & ECCV'18 & $256\times128$ & 93.3 & 80.9 \\
OSNet \cite{zhou2019omni} & ICCV'19 & $256\times128$ & 94.8 & 84.9 \\
ISP \cite{zhu2020identity} & ECCV'20 & $256\times128$ & 95.3 & 88.6 \\
CDNet \cite{li2021combined} & CVPR'21 & $256\times128$ & 95.1 & 86.0 \\ 
FSAM \cite{hong2021fine} & CVPR'21 & $256\times128$ & 94.6 & 85.6 \\
CAL \cite{gu2022clothes} & CVPR'22 & $384\times192$ & 94.7 & 87.5 \\
IMS-GEP \cite{zhao2023joint}& TMM'23 & $384\times192$ & 95.1 & 87.1 \\
AFL \cite{liu2023clothes} & TMM'23 & $384\times192$ & 95.5 & 88.8 \\ \hline
Ours & - & $256\times128$ & 96.0 & 90.2 \\
Ours & - & $384\times192$ & \textbf{96.5} & \textbf{91.4} \\ \hline
\end{tabular}
\label{tab:market}
\end{center}
\end{table}

\subsubsection{Effect of CDA}
In order to verify the effectiveness of our proposed clothes diversity augmentation (CDA) for the \ccreid task, we conduct ablation experiments on the LTCC and PRCC datasets.
The experimental results are shown in Table \ref{tab:ablation}, which shows that the proposed CDA can bring significant performance improvements.
For example, on the LTCC dataset, Method 3 outperforms Method 1 (baseline) by 5.0\% Rank-1 and 4.4\% mAP.
In addition, we compare the CDA with the DA \cite{chen2023unsupervised} mentioned in Section \ref{sec:3}.B.
It can be observed that Method 3 achieves superior performance to Method 2 on both LTCC and PRCC datasets.
The reason is that CDA only changes the pixels in the clothing area, avoiding the negative impact on model training caused by appearance changes in identity-related regions such as the head and skin.

\subsubsection{Effect of MCB}
\begin{figure}[t]
  \centering
  \includegraphics[width=\linewidth]{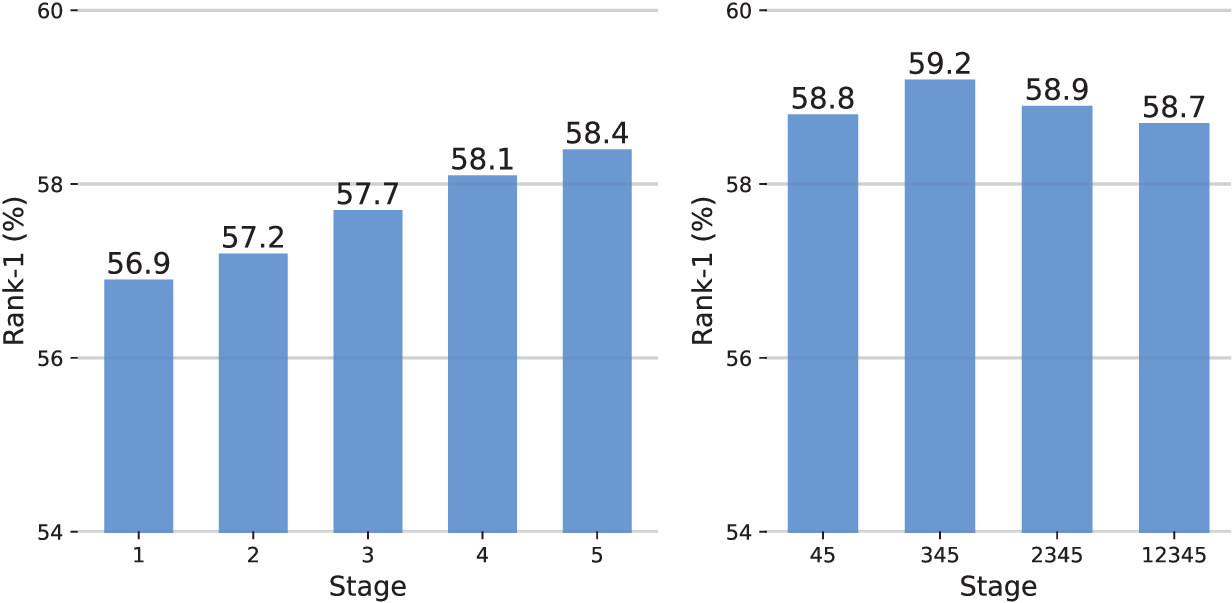}
  \caption{Some experimental results on the insertion position of the MCB. Left: The impact of inserting MCB after each stage of the backbone. Right: The impact of adding MCB forward from the last stage of the backbone.}
\label{fig:MCB}
\end{figure}

According to the performance of Method 3 and Method 4 in Table \ref{tab:ablation}, our proposed multi-scale constraint block (MCB) achieves 4.1\% Rank-1 and 4.8\% mAP improvements on the LTCC dataset.
Similarly, on the PRCC dataset, MCB obtains 2.8\% Rank-1 and 2.4\% mAP improvements.
This indicates that MCB can fully learn fine-grained identity information and achieve cloth-agnostic knowledge transfer, increasing the resistance of the model to clothing variations.
Given a standard ResNet50 with 5 stages, we further explore the most suitable location for the insertion of the MCB, and a few experimental results are displayed in Fig. \ref{fig:MCB}.
This experiment is performed using the PRCC dataset under the cloth-changing setting, with the results being reported in terms of Rank-1 accuracy.
It can be seen from the results that (1) when inserting MCB after each stage of the backbone, stage 5 achieves the best performance. (2) When more MCBs are inserted forward from stage 5, stages 3, 4 and 5 can obtain the best results, but the performance starts to decrease when too many blocks are inserted.
It might be because the shallow layers of ResNet50 mainly extract edge and texture features, while the deep layers of ResNet50 learn more identity-related information.
Aligning high-level identity features is more beneficial to the person \reid.

\begin{table*}[t]
\caption{Ablation experiments of the proposed \modelnameshort on the LTCC and PRCC datasets. "DA", "CDA", "MCB", "CAM", "CI" and "SAC" denote the domain augmentation, clothes diversity augmentation, multi-scale constraint block, counterfactual-guided attention module, counterfactual intervention and semantic alignment constraint, respectively.}
\setlength{\tabcolsep}{3mm}
\renewcommand{\arraystretch}{1.1}
\centering
\begin{tabular}{ccccccccccc}
\cline{1-11}
\multicolumn{1}{c|}{\multirow{2}{*}{Method}} & \multirow{2}{*}{DA} & \multirow{2}{*}{CDA} & \multirow{2}{*}{MCB} & \multirow{2}{*}{CAM} & \multirow{2}{*}{CI} & \multicolumn{1}{c|}{\multirow{2}{*}{SAC}} & \multicolumn{2}{c|}{LTCC} & \multicolumn{2}{c}{PRCC} \\ \cline{8-11}
\multicolumn{1}{c|}{} &  &  &  &  &  & \multicolumn{1}{c|}{} & Rank-1 & \multicolumn{1}{c|}{mAP} & Rank-1 & mAP\\ \hline
\multicolumn{1}{c|}{1 (baseline)} & $\times$ & $\times$ & $\times$ & $\times$ & $\times$ & \multicolumn{1}{c|}{$\times$} & 38.3 & \multicolumn{1}{c|}{19.1} & 52.5 & 50.8 \\ 
\multicolumn{1}{c|}{2} & \checked & $\times$ & $\times$ & $\times$ & $\times$ & \multicolumn{1}{c|}{$\times$} & 40.2 & \multicolumn{1}{c|}{20.7} & 53.9 & 52.0 \\ 
\multicolumn{1}{c|}{3} & $\times$ & \checked & $\times$ & $\times$ & $\times$ & \multicolumn{1}{c|}{$\times$} & 43.3 & \multicolumn{1}{c|}{23.5} & 56.4 & 54.6 \\ 
\multicolumn{1}{c|}{4} & $\times$ & \checked & \checked & $\times$ & $\times$ & \multicolumn{1}{c|}{$\times$} & 47.4 & \multicolumn{1}{c|}{28.3} & 59.2 & 57.0 \\ 
\multicolumn{1}{c|}{5} & $\times$ & \checked & \checked & \checked & $\times$ & \multicolumn{1}{c|}{$\times$} & 48.2 & \multicolumn{1}{c|}{30.4} & 60.6 & 58.3 \\ 
\multicolumn{1}{c|}{6} & $\times$ & \checked & \checked & \checked & \checked & \multicolumn{1}{c|}{$\times$} & 49.5 & \multicolumn{1}{c|}{33.0} & 61.8 & 60.5 \\
\multicolumn{1}{c|}{7} & $\times$ & \checked & \checked & \checked & \checked & \multicolumn{1}{c|}{\checked} & 53.1 & \multicolumn{1}{c|}{35.9} & 64.9 & 64.6 \\ \hline
\end{tabular}
\label{tab:ablation}
\end{table*}

\subsubsection{Effect of CAM}
From the results of Method 4 and Method 6 in Table \ref{tab:ablation}, it can be seen that the addition of our counterfactual-guided attention module (CAM) brings significant performance improvement to the model.
In particular, the addition of CAM improves the model by 4.7\% mAP on the LTCC dataset.
This indicates that CAM can effectively help the model extract cloth-agnostic features.
In addition, we explore the impact of counterfactual intervention.
When counterfactual intervention is not used to supervise attention learning (Method 5), the mAP performance on the LTCC and PRCC datasets decreases by 2.6\% and 2.2\%, respectively.
This demonstrates the effectiveness of applying counterfactual intervention to the model.

\subsubsection{Effect of SAC}
As shown in Table \ref{tab:ablation}, compared with Method 6, our designed semantic alignment constraint (SAC) can significantly improve model performance on both datasets. 
Specifically, it improves 3.6\% Rank-1 and 2.9\% mAP on the LTCC dataset, while on the PRCC dataset, SAC improves 3.1\% Rank-1 and 4.1\% mAP.
This suggests that SAC can effectively achieve the transfer of identity-related semantic information, further enhancing the ability of the model to distinguish pedestrian identities.

\begin{figure}[t]
  \centering
  \includegraphics[width=\linewidth,height=9cm]{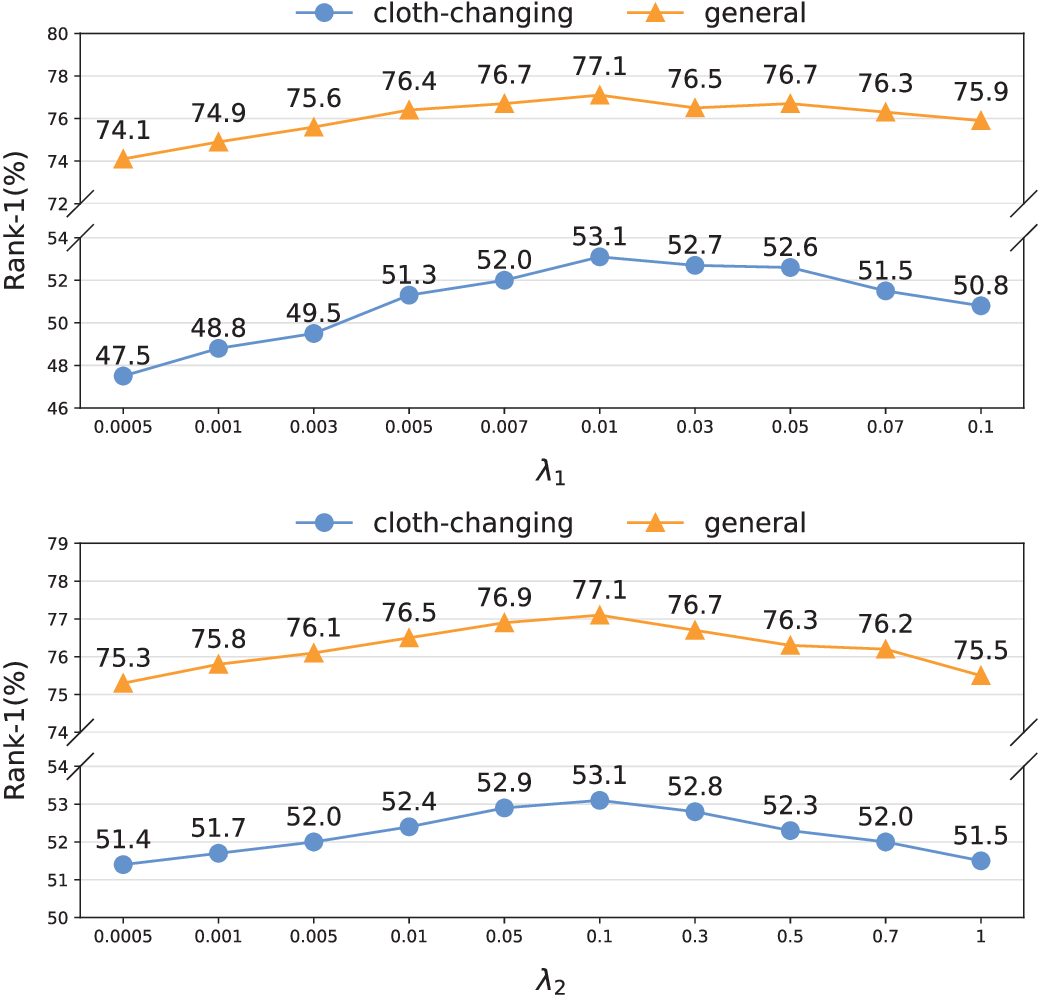}
  \caption{Ablation experiments of hyper-parameters. Above: The effect of $\lambda_1$. Below: The effect of $\lambda_2$.}
  \label{fig:lambda}
\end{figure}

\subsubsection{Effect of Hyper-parameters}
In Eq. \eqref{eq:total}, we introduce two hyper-parameters $\lambda_1$ and $\lambda_2$ to adjust the weight of loss functions.
The experiment is conducted using the LTCC dataset in general and cloth-changing scenarios to evaluate the impact of these hyper-parameters, and results are shown in Fig. \ref{fig:lambda}.
When testing one hyper-parameter, we maintain the other at its optimal value.
It can be observed that when $\lambda_1$ equals 0.01 and $\lambda_2$ equals 0.1, the model achieves the best performance on the LTCC dataset.
Moreover, when $\lambda_1$ and $\lambda_2$ increase, the performance of \modelnameshort constantly improves, while it begins to degrade when $\lambda_1$ and $\lambda_2$ exceed 0.01 and 0.1, respectively.
Therefore, selecting 0.01 for $\lambda_1$ and 0.1 for $\lambda_2$ is reasonable.
It might be because moderate knowledge transfer enables the model to adaptively learn appearance information and cloth-irrelevant information. 
However, imposing stronger constraints on the attention module can help the model adequately extract discriminative cloth-agnostic features.

\subsection{Complexity Analysis}
We compare the parameters and running time of our proposed \modelnameshort with other state-of-the-art methods on the PRCC dataset, as shown in Table \ref{tab:complexity}.
This experiment follows the experimental environment in \cite{yang2023good} for a fair comparison and is performed under the cloth-changing setting.
It can be observed that the proposed method has competitive parameters and running time in both the training and testing stages and achieves better performance than other methods.
In the testing stage, the \modelnameshort discards the clothing erasing stream and does not rely on human parsing and data augmentation, further reducing the parameters and testing time.
Therefore, the proposed method is efficient for the \ccreid task.


\begin{table}[t]
\caption{Comparison of network parameters (Params) and running time with state-of-the-art methods.}
\vspace{-2mm}
\begin{center}
\setlength{\tabcolsep}{1.7mm}
\renewcommand{\arraystretch}{1.1}
\begin{tabular}{c|c|cc|cc|c}
\hline
\multirow{2}{*}{Method} & \multirow{2}{*}{Venue} & \multicolumn{2}{c|}{Training} & \multicolumn{2}{c|}{Testing} & PRCC \\ \cline{3-7} 
& & Params & Time & Params & Time & Rank-1 \\ \hline
RGA \cite{zhang2020relation} & CVPR'20 & 30.13M & 0.8h & 30.13M & 40s & 42.3 \\ 
ISP \cite{zhu2020identity} & ECCV'20 & 31.68M & 16.5h & 31.68M & 30s & 36.6 \\
FSAM \cite{hong2021fine} & CVPR'21 & 164.27M & 12h & 23.82M & 15s & 54.5 \\
CAL \cite{gu2022clothes} & CVPR'22 & 23.52M & 2.2h & 23.52M & 58s & 54.6 \\
AIM \cite{yang2023good} & CVPR'23 & 72.67M & 4.1h & 23.52M & 58s & 57.8 \\ \hline
Ours & - & 27.89M & 2.5h & 25.08M & 27s & 64.4 \\ \hline
\end{tabular}
\end{center}
\label{tab:complexity}
\end{table}

\begin{figure}[t]
  \centering
  \includegraphics[width=\linewidth, height=3.8cm]{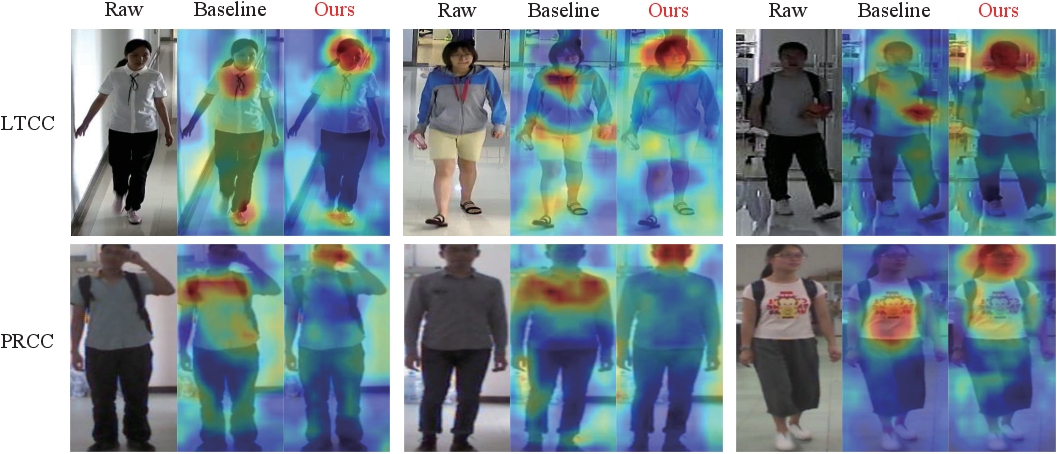}
  \caption{Raw images and activation maps generated by baseline and \modelnameshort.}
  \label{fig:activation}
\end{figure}
\begin{figure}[t]
  \centering
  \includegraphics[width=\linewidth,height=2.9cm]{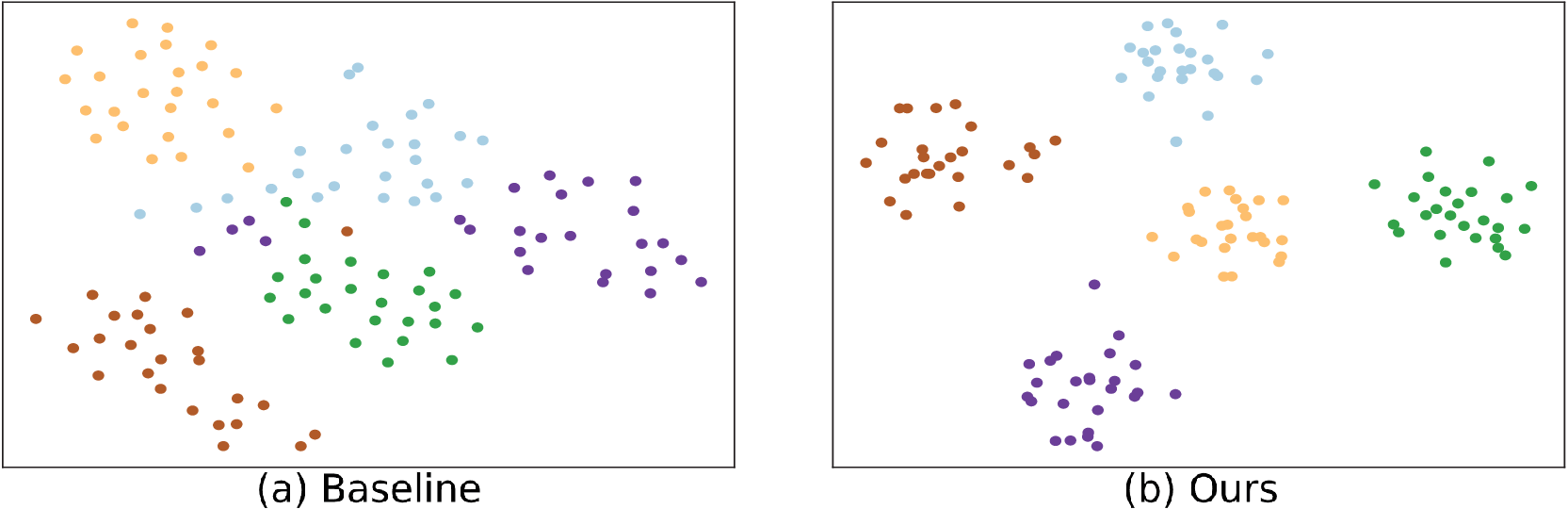}
  \caption{The t-SNE visualization of features on the PRCC dataset. We randomly choose 5 identities, corresponding to specific colors in the figure.}
  \label{fig:tsne}
\end{figure}
\begin{figure}[t]
  \centering
  \includegraphics[width=\linewidth,height=6.5cm]{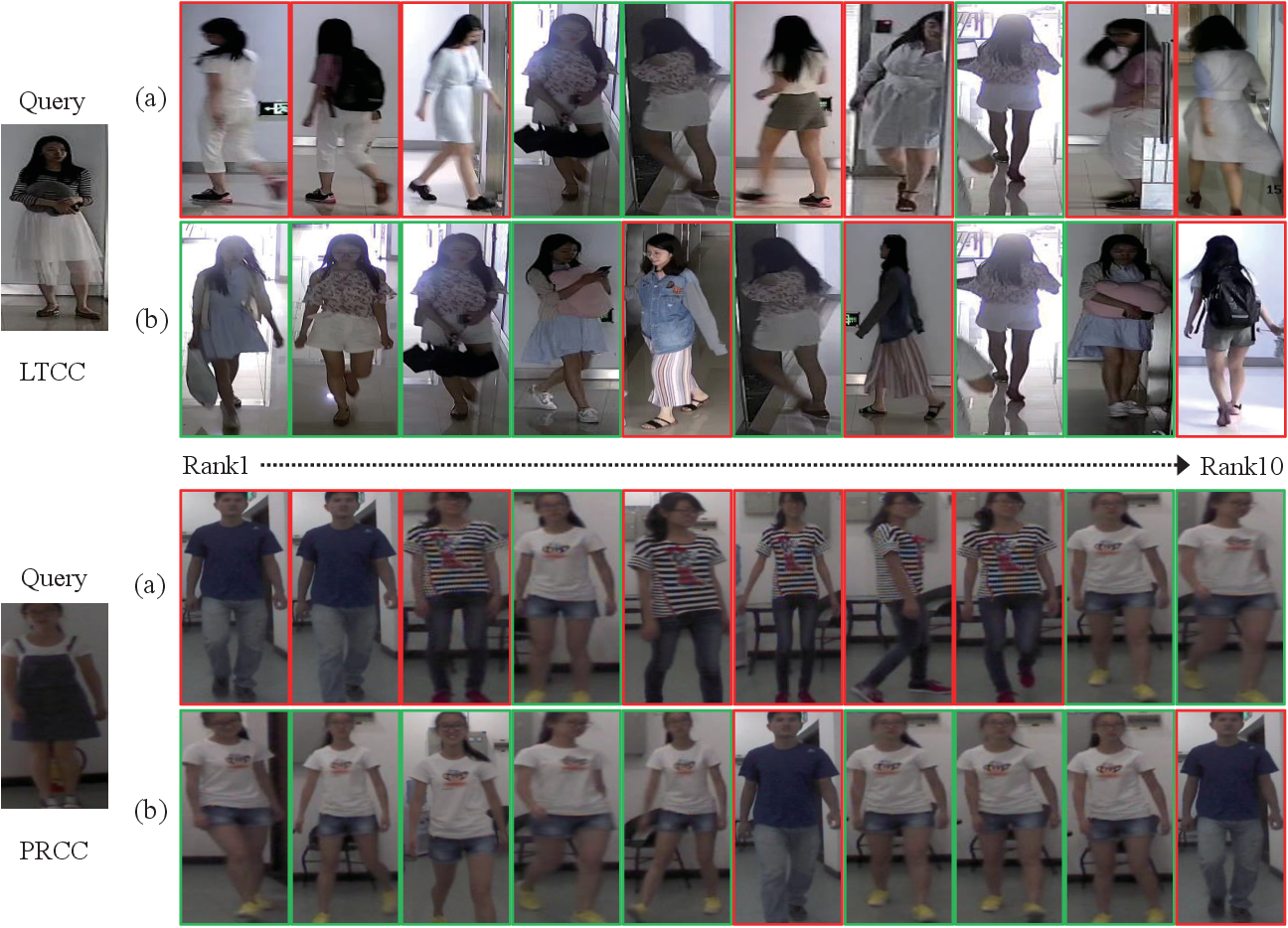}
  \caption{Visualization of the top-10 retrieval results from baseline and \modelnameshort on the LTCC and PRCC datasets. The green and red boxes show positive and negative results. (a) Ranking lists of baseline. (b) Ranking lists of \modelnameshort.}
  \label{fig:ranking}
\end{figure}

\subsection{Visualization Analysis} \label{sec:visual}
\subsubsection{Activation Maps}
To explore the clues learned by the model, we generate the activation maps of both baseline and proposed \modelnameshort, and the results on the LTCC and PRCC datasets are shown in Fig. \ref{fig:activation}.
The activation maps can visually indicate the regions that receive greater attention from the model.
It can be seen that the activation maps of the baseline show scattered highlights.
In contrast, the activation maps of \modelnameshort show more centralized highlights, which indicates that the \modelnameshort suffers from more constraints during the training process.
Moreover, compared with the baseline that primarily focuses on cloth-relevant areas such as the upper clothes, the proposed \modelnameshort focuses more on cloth-agnostic areas such as the head.
This demonstrates that the proposed method can fully explore cloth-irrelevant clues and has better robustness to cloth-changing situations.
Notably, our proposed method can maintain precise attention to regions unrelated to clothing despite the lower resolution of the PRCC dataset, which shows its effectiveness in intelligent surveillance applications.

\subsubsection{Feature Distributions}
In order to research the distribution of features learned by the model, we conduct the t-SNE \cite{van2008visualizing} visualization experiment on the testing set of PRCC, which compares the feature distribution in the latent space of the baseline and proposed \modelnameshort, as shown in Fig. \ref{fig:tsne}.
For the baseline, it can be observed that the features within each category exhibit discretization, and features from different categories are confused together.
In contrast, the features learned by \modelnameshort have better intra-class compactness and inter-class discrepancy.
This indicates that our method can effectively learn discriminative identity-related features and mitigate the interference caused by clothing variations.

\subsubsection{Retrieval Results}
To provide an intuitive evaluation of the proposed \modelnameshort, we present visualizations of the top-10 retrieval results on the LTCC and PRCC datasets for both the baseline network and our \modelnameshort, and results are shown in Fig. \ref{fig:ranking}.
This experiment is conducted under the cloth-changing setting, where gallery samples sharing the same identity and clothing as the query are excluded.
It can be seen that the LTCC dataset presents more significant challenges in terms of variations in clothing, illumination and gestures.
On both datasets, the proposed method shows better retrieval performance than the baseline model.
For instance, the retrieval result in the first line shows that the baseline is susceptible to clothing color, leading to incorrect matches with persons dressed in white skirts or white shorts.
The retrieval result in the third line indicates that clothing texture disrupts the baseline, resulting in mismatches with pedestrians wearing striped clothes.
Compared with the baseline, our proposed method effectively mitigates the interference from clothing color and texture and achieves superior retrieval quality.

\section{Conclusion}
In this paper, a novel identity-aware dual-constraint network is proposed for the \ccreid task, which consists of the raw image stream and clothing erasing stream.
For insufficient cloth-changing samples in existing \ccreid datasets, a clothes diversity augmentation is presented, which generates more realistic cloth-changing pedestrian images by enriching the color of clothes while preserving the texture.
To achieve cloth-irrelevant knowledge transfer, we design a multi-scale constraint block, which utilizes the proposed hierarchical matching loss to conduct feature alignment of two streams in the backbone network.
Moreover, a counterfactual-guided attention module is introduced, which learns cloth-agnostic clues from channel and spatial dimensions and leverages the proposed cloth-agnostic contrastive loss to perform the counterfactual intervention.
Finally, a semantic alignment constraint is designed to effectively facilitate the raw image stream to learn cloth-irrelevant semantic features.
The framework is trained end-to-end, while the clothing erasing stream is discarded during the testing stage.
Extensive experiments on several public general \reid and \ccreid datasets demonstrate that our \modelnameshort outperforms state-of-the-art methods. \par
The precision of the human parsing model partially limits the effectiveness of our proposed method.
In future work, we aim to reduce the dependence of the proposed approach on the auxiliary human parsing model.
Moreover, we will extend our ideas to other tasks, such as object detection, vehicle re-identification and face recognition.

\bibliographystyle{IEEEtran}
\bibliography{ref}

\vfill

\end{document}